\documentclass[letterpaper, 10 pt, conference]{ieeeconf}  % Comment this line out if you need a4paper

\IEEEoverridecommandlockouts                              % This command is only needed if 
\usepackage{amsmath,amssymb,amsfonts}
\usepackage{graphicx}
\usepackage{algorithmic}
\usepackage[ruled,vlined]{algorithm2e}
\usepackage{cite}
\usepackage{textcomp}
\usepackage{xcolor}
\usepackage{pifont}
\usepackage[percent]{overpic}
\usepackage{hyperref}
\usepackage[utf8]{inputenc}
\usepackage{setspace}
\title{\LARGE \bf
TacUMI: A Multi-Modal Universal Manipulation Interface for Contact-Rich Tasks\\
}

\author{Tailai Cheng$^{1,2}$, Kejia Chen$^{1}$, Lingyun Chen$^{1}$, Liding Zhang$^{1}$, Yue Zhang$^{1}$, Yao Ling$^{1}$,\\  Mahdi Hamad$^{2}$, Zhenshan Bing$^{3,1}$, Fan Wu$^{4,1}$, Karan Sharma$^{2}$, Alois Knoll$^{1}$% <-this % stops a space
\thanks{$^{1}$ School of Computation, Information and Technology, Technical University of Munich, Germany.}
\thanks{$^{2}$Agile Robots SE, Munich, Germany}%}%
\thanks{$^{3}$ State Key Laboratory for Novel Software Technology and the School of Science and Technology, Nanjing University (Suzhou Campus), China.}
\thanks{$^{4}$Shanghai University, China.}
}

\begin{document}
\maketitle
% \thispagestyle{empty}
% \pagestyle{empty}

%%%%%%%%%%%%%%%%%%%%%%%%%%%%%%%%%%%%%%%%%%%%%%%%%%%%%%%%%%%%%%%%%%%%%%%%%%%%%%%%
\begin{abstract}

Task decomposition is critical for understanding and learning complex long-horizon manipulation tasks.
Especially for tasks involving rich physical interactions, relying solely on visual observations and robot proprioceptive information often fails to reveal the underlying event transitions. 
% In such scenarios, where multiple fine-grained skills are sequentially composed, identifying event boundaries provides an effective way to decompose demonstrations into meaningful units. 
% This highlights two key requirements: fast collection of high-quality multimodal data and robust segmentation of long sequential tasks.
This raises the requirement for efficient collection of high-quality multi-modal data as well as robust segmentation method to decompose demonstrations into meaningful modules.

Building on the idea of the handheld demonstration device Universal Manipulation Interface (UMI), we introduce TacUMI, a multi-modal data collection system that integrates additionally ViTac sensors, force–torque sensor, and pose tracker into a compact, robot-compatible gripper design, which enables synchronized acquisition of all these modalities during human demonstrations.
% , which provides abundant information for understanding fine-grained tasks. 
We then propose a multi-modal segmentation framework that leverages temporal models to detect semantically meaningful event boundaries in sequential manipulations.
Evaluation on a challenging cable mounting task shows more than 90\% segmentation accuracy and highlights a remarkable improvement with more modalities, which validates that TacUMI establishes a practical foundation for both scalable collection and segmentation of multi-modal demonstrations in contact-rich tasks.
The design and experiment results are available at our project website:~\href{https://tac-umi.github.io/TacUMI/}{https://tac-umi.github.io/TacUMI/}.
% Our system is validated on a challenging cable mounting task, 
% showing that multi-modal sensing substantially improves segmentation accuracy and reveals transferable skill structures across platforms. 
% Together, TacUMI establishes a practical foundation for scalable collection and segmentation of multi-modal demonstrations in contact-rich manipulation.

\end{abstract}

%%%%%%%%%%%%%%%%%%%%%%%%%%%%%%%%%%%%%%%%%%%%%%%%%%%%%%%%%%%%%%%%%%%%%%%%%%%%%%%%
\section{INTRODUCTION}
% This document is a model and instructions for \LaTeX.
% Please observe the conference page limits. 

In real-world applications such as industrial assembly and household services, robots are often required to perform long-horizon tasks consisting of a sequence of modular skills. 
% Decomposing these complex manipulation tasks into semantically meaningful short events is thus critical for robots to understand and learn from them.
% Segmenting and understanding complex manipulation tasks remains a fundamental challenge in robotic skill learning.
However, learning such long-horizon tasks, especially when there are intensive physical interactions, remains a fundamental challenge.
While recent imitation learning frameworks, such as ACT~\cite{ACT} and Diffusion Policy~\cite{diffusion_policy}, have achieved strong performance on short-horizon tasks, they treat the whole demonstration as a monolithic sequence without explicit segmentation, which makes them struggle to learn long-horizon tasks at once~\cite{diffusion_survey}.
In contrast, learning separately modular motor skills and high-level coordinator to compose them have proven to be an efficient and promising paradigm~\cite{Bottom-Up, task_networks}.

\begin{figure}[thpb]
    \centering  
    \includegraphics[width=0.95\columnwidth]{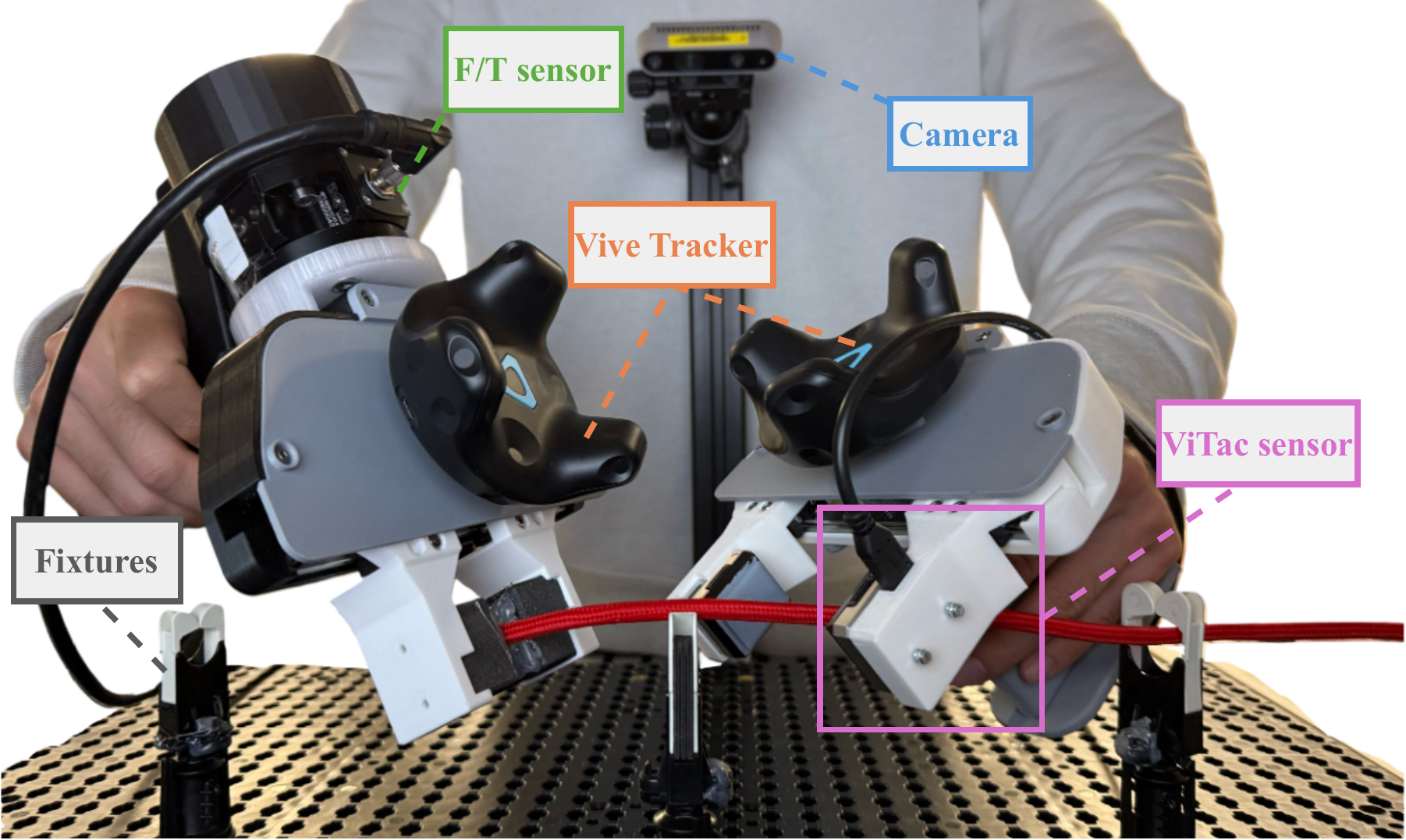}
    \caption{Overall setup for the cable mounting task. 
    The left hand holds a modified UMI gripper equipped with a ViTac sensor and a Vive Tracker, while the right hand holds the proposed TacUMI gripper for multimodal data collection. 
    A camera positioned in the center records third-person visual information during the demonstrations.}
    \label{fig:whole_setup}
    \vspace{-1.0em}
\end{figure}

This hierarchical learning paradigm requires firstly decomposing a long demonstration into semantically meaningful, modular skills.
In this segmentation, tactile information can play an important role, as it can capture contact dynamics that may be missed by visual observation~\cite{LEMMo-Plan}.
For example, during cable mounting, operators often stretch the cable to create tension.
This step is difficult to discern visually but can be clearly captured through tactile signals.
However, tactile sensing has not been sufficiently explored in existing research due to limitation in both data acquisition and learning methods.

% current technologies face significant limitations in both data acquisition methods and learning algorithms.

In terms of data collection, teleoperation-based systems~\cite{VideoDex,VIOLA,learningwholebodyhumanrobothaptic} provide precise demonstrations, but their high cost, lack of intuitive control, and limited force feedback severely hinder practical usability.
Recent handheld data collection devices, such as the Universal Manipulation Interface (UMI)~\cite{umi} and FastUMI~\cite{FastUMI}, have lowered the barrier for human-in-the-loop demonstration.
Nevertheless, they rely solely on vision and pose, and the critical tactile information is missing from their design.
% yet their reliance on solely visual modality and pose data restricts the acquisition of critical tactile information, 
When it comes to learning methods, visual and proprioceptive data have been widely explored for task decomposition~\cite{Bottom-Up, XSkill, SKIDRAW}.
However, less attention were paid to the utilization and integration of tactile sensing.
% but for sequential, contact-intensive tasks, tactile information is crucial for capturing rich dynamics that can be missed by visual observations.

% As a result, they struggle to capture task-specific skills in contact-rich settings and cannot easily reorganize or reuse these skills for new task configurations.
% In particular, manipulation tasks involving Deformable Linear Objects (DLOs)—such as cable routing—pose unique challenges. 
% Visual and pose data alone are insufficient to infer crucial physical states like tension or stiffness, which are essential for skills such as \textit{pulling taut}.
% Moreover, in long-horizon, contact-rich tasks, solving the problem requires not only recognizing fine-grained transitions but also the ability to \textit{reorder and reuse skills} across different task configurations. 
% Without explicit task segmentation, learning remains monolithic, preventing the decomposition of demonstrations into transferable skills and thus limiting generalization to unseen scenes.

To address these limitations, we present a new solution that tightly integrates multi-modal sensing hardware with a task segmentation learning framework. 
On the hardware side, we develop an handheld data collection device inspired by UMI, but enhanced with ViTac sensors on fingertips, a 6D force-torque (F/T) sensor on the wrist, and a high-precision 6D pose tracker.
Especially for F/T signals, we introduce a continuously lockable jaw mechanism that eliminates the user’s interference when grasping.
This enables synchronize collecting clean measurement of contact tactile, forces/torque, and gripper pose information during demonstrations.

On the algorithmic side, we propose a BiLSTM-based task segmentation framework that learns to partition long-horizon demonstrations into a sequence of skills. 
Using the rich temporal structure and complementary signals from multiple modalities, our approach identifies transition boundaries between each stage of long-horizon manipulation tasks. 
This segmentation lays a foundation for learning modular skills and composing them, which makes learning from long-horizon tasks more scalable and interpretable.
% improving scalability and interpretability.

To validate our proposed data collection and segmentation approach, we evaluate our system on a challenging cable mounting task.
In comparison to other UMI-based designs as well as teleoperation, TacUMI obtains uninterferred F/T data in the most efficient way.
% where the multi-modal data collected by our custom-designed gripper enables reliable task segmentation through a BiLSTM-based model. 
By integrating multiple, especially the tactile, modalities collected by TacUMI, the learned segmentation model reveals clear and semantically meaningful transition across different phases of the manipulation process.
Moreover, we deploy the model trained from TacUMI-collected dataset on another dataset collected by teleoperating robots, and achieves comparable segmentation accuracy.
These evaluations demonstrate the effectiveness of our system for interpreting and segmenting long-horizon manipulation tasks.

% \begin{itemize}
%     \item \textbf{Multimodal Sensing Integration:}
%     Our system fuses data from fingertip-mounted ViTac sensors, a 6D force-torque sensor on the wrist, and a high-precision 6D pose tracker. This enables accurate, synchronized measurement of contact tactile, forces/torque, and gripper pose information during demonstrations.
    
%     \item \textbf{Interference-Free Mechanical Design:}
%     We introduce a continuously lockable jaw mechanism that eliminates the user’s actuation force from being captured by the force sensor, ensuring clean force measurements during grasping.
    
%     \item \textbf{Universal Grasping Adaptability:} 
%     The gripper can exert arbitrarily large gripping forces to securely hold objects of diverse shapes and sizes, without requiring part replacement or task-specific adjustments.
    
%     \item \textbf{Robust and Efficient Pose Tracking:}
%     By using a 6D pose tracker for pose estimation, our system avoids the drift accumulation commonly associated with vision-based SLAM systems.
% \end{itemize}
\section{RELATED WORK}
\subsection{Data Collection Interfaces for Robot}
High-quality multimodal data is crucial for understanding complex, contact-rich manipulation tasks.
Conventional teleoperation systems~\cite{GELLO, Bridge_data, diffusion_policy, LEMMo-Plan} offer precise control and direct mapping to robot embodiments, but they are often costly, cumbersome to set up, and lack intuitive force feedback for the operator. 
These limitations have motivated the development of \textit{handheld robotic data collection devices} that allow operators to physically manipulate objects in a natural manner while capturing sensor data.

% The \textbf{Universal Manipulation Interface (UMI)}\cite{umi} introduced a low-cost handheld device equipped with a monocular camera and ORB-SLAM3-based visual-inertial tracking, enabling collection of visual and pose data from human demonstrations. 
% While UMI reduces hardware complexity, its reliance on vision-only sensing limits the ability to capture fine-grained contact information, and pose estimation can suffer from drift in long-horizon tasks. 
% \textbf{FastUMI}\cite{FastUMI} improves hardware portability and enhances SLAM stability by using an Intel RealSense T265 for pose estimation. Their approach additionally incorporates periodic returns to a predefined reference position during demonstrations to recalibrate and mitigate cumulative pose drift, but still depends on vision-only modalities and does not address force sensing or tactile perception. 
% \textbf{ForceMimic}~\cite{ForceMimic} incorporates a 6D force-torque (FT) sensor and a ratchet-based locking mechanism, enabling capture of wrench data during manipulation. However, its soft gripper design and discrete locking mechanism limits maximum clamping force, leading to slippage in deformable linear object (DLO) manipulation.
The Universal Manipulation Interface (UMI)\cite{umi} and FastUMI\cite{FastUMI} provide low-cost handheld tools that enable rapid collection of visual and pose data from human demonstrations.
While effective for efficient data acquisition, both systems rely solely on vision-based sensing and thus cannot capture tactile feedback or the interaction forces between the gripper and the environment.
Moreover, their pose estimation, built on SLAM-based methods, inevitably suffers from cumulative drift during long-horizon demonstrations.
ForceMimic~\cite{ForceMimic} extends data collection by incorporating a 6D force-torque (F/T) sensor and a ratchet-based locking mechanism, allowing the recording of wrench data during manipulation. However, its soft gripper structure and discrete locking mechanism limit the achievable clamping force and stability. 
As a result, the gripper can fail to provide sufficient normal force and friction at the contact interface, leading to slippage or even loss of the manipulated object under external loads.

In contrast, our device integrates synchronized \textit{tactile sensing}, \textit{force-torque measurement}, and \textit{precise 6-DoF pose tracking} via a pose tracker, eliminating cumulative drift issues in long tasks. 
The continuous locking mechanism allows the operator to maintain a stable grasp without sustaining trigger pressure, removing internal actuation forces from FT recordings and ensuring clean interaction data. 
Furthermore, our gripper’s high clamping force capability prevents slippage during high-tension interactions (e.g., DLO manipulation), extending applicability to a wider range of contact-rich scenarios.

\subsection{Event Segmentation}
Event segmentation aims to decompose continuous demonstrations into semantically meaningful skills, which is essential for learning and executing long-horizon tasks.  
Unsupervised methods such as Bottom-Up Skill Discovery~\cite{Bottom-Up} employ agglomerative clustering to build hierarchical task structures from raw demonstrations, identifying recurring motion patterns as candidate skills. 
XSkill~\cite{XSkill} learns a shared skill representation space and a set of learnable skill prototypes from unlabelled human and robot videos, aligning them into a unified latent space for cross-embodiment skill discovery. 
SkIDRAW~\cite{SKIDRAW} jointly performs unsupervised skill discovery and trajectory segmentation directly from raw, unlabelled trajectories. 
While these approaches are effective for discovering latent skills without annotations, they are unsuitable for our cable mounting task, where each skill is explicitly defined and unsupervised clustering may lead to ambiguous or inconsistent segmentation.

Other segmentation works, such as \cite{Online_task_segmentation,temporal_segmentation_human_hand_motion}, operate by detecting motion boundaries from human or robot demonstrations, relying primarily on camera video and proprioceptive data. However, these modalities alone are insufficient for our scenario: in cable mounting, certain skills, such as \textit{tensioning the cable}, involve minimal observable motion and thus cannot be reliably detected using only visual and proprioceptive cues.

Our segmentation approach follows the \textit{object-centric} perspective from LEMMo-Plan~\cite{LEMMo-Plan}, where segmentation is driven by changes in the state of the manipulated object rather than solely by end-effector kinematics. 
Although LEMMo-Plan showed that ViTac images can directly capture subtle skill transitions during object manipulation, we found that using tactile data alone was insufficient to achieve robust performance. 
% we found that using a TimeSformer-based\cite{timesformer} segmentation model on tactile data alone was insufficient for robust performance. 
To address this, we integrate ViTac data, force-torque measurements, and third-person visual observations, and process them using a BiLSTM-based temporal model. 
This multi-modal integration enables reliable detection of subtle contact-state changes and accurate segmentation of complex, long-horizon cable mounting demonstrations.

\section{METHODOLOGY}

\subsection{Mechanical Design for Robot-Executable Demonstrations}\label{sec:hardware_design}
\begin{figure}[thpb]
    \centering  
    \includegraphics[width=0.95\columnwidth]{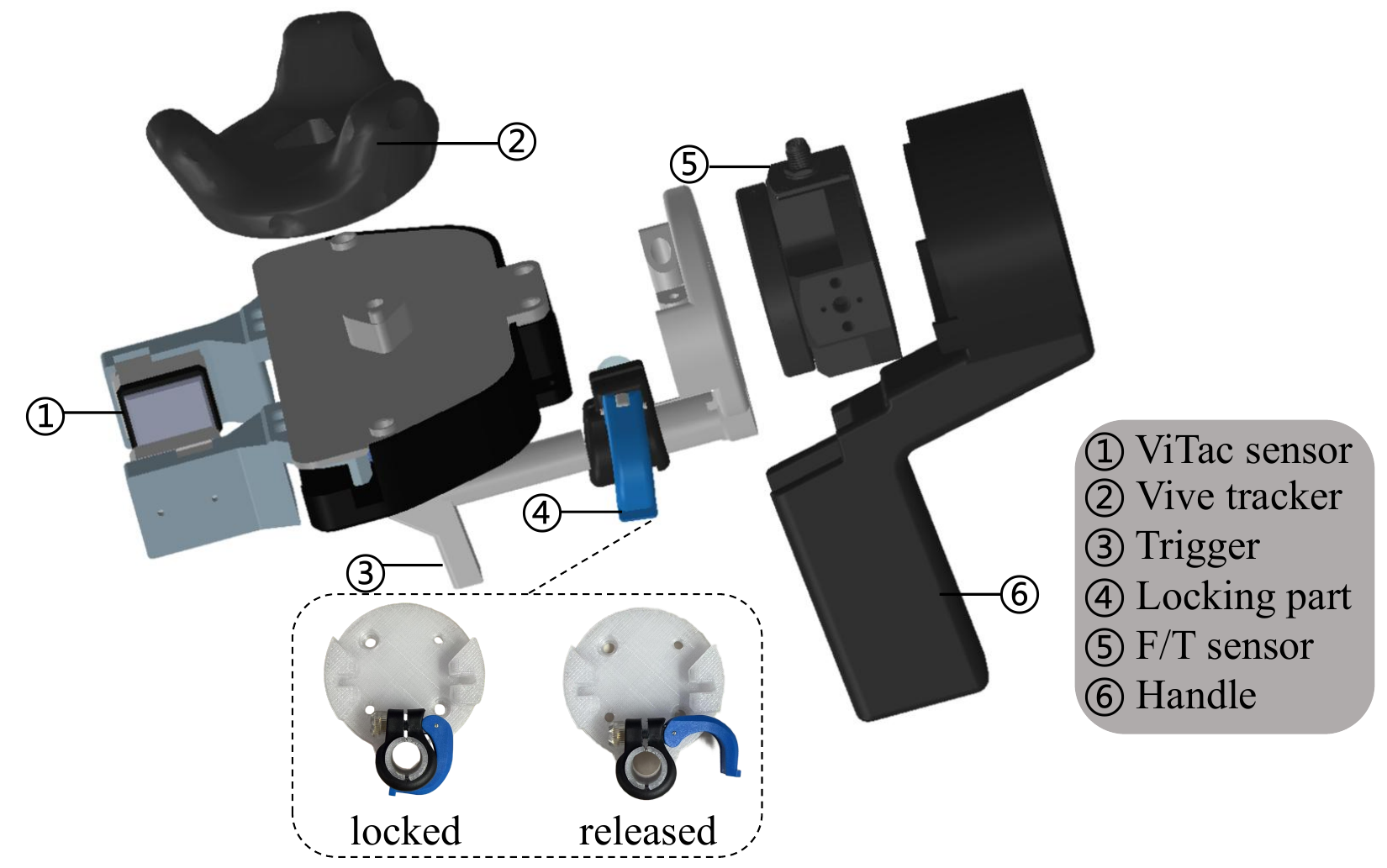}
    \caption{TacUMI structure.
    Exploded view of the proposed multimodal data collection gripper, showing: 
    (1) ViTac sensor (GelSight Mini); 
    (2) Vive tracker for 6-DoF pose tracking; 
    (3) Trigger connected to the rack-and-pinion mechanism for grasp actuation; 
    (4) Continuous locking part that enables the trigger to be locked at arbitrary jaw opening widths; 
    (5) {6-axis Force/Torque (F/T) sensor}; 
    (6) {Handle} ergonomically designed for single-handed operation. 
    Insets illustrate the locking mechanism in the locked and released states.}
    \label{fig:gripper_structure}
    \vspace{-1em}
\end{figure}

We propose a new handheld demonstration device that supports multi-modal, synchronized, and robot-compatible data collection for complex manipulation tasks. 
Inspired by previous designs\cite{umi, FastUMI, ForceMimic}, our system significantly extends hardware capabilities for demonstrating long-horizon and contact-rich manipulation tasks.
The mechanism preserves a robot wrist–end effector layout and integrates tactile sensing, 6D force–torque (F/T) sensing, and 6D pose tracking in a compact, one-handed form factor (Fig.~\ref{fig:gripper_structure})

\subsubsection{Robot-Compatible Force–Torque Stack}
A 6-axis F/T sensor is mounted between the gripper and a rear handle shaped as a \textit{standard robot flange} (Fig.~\ref{fig:gripper_structure} \textcircled{5}). 
This preserves the physical stack-up used on robots, so the measured wrench during human operation matches robot–environment interactions at execution. 
Because the F/T sensor sits between the wrist-equivalent interface and the tool, the recorded wrench is directly transferable to downstream control without frame re-ordering or ad-hoc calibration.

\subsubsection{Rack-and-Pinion Trigger with Continuous Self-Locking}
We retain the UMI-style rack-and-pinion trigger for intuitive force scaling, where finger closure is proportional to trigger effort~\cite{umi}. 
In prior designs, continuous trigger pressing was required to maintain a grasp, and this actuation leaked into F/T measurements.
To avoid this, we introduce a \textbf{continuous self-locking mechanism} (Fig.~\ref{fig:gripper_structure} \textcircled{4}) that rigidly fixes the jaw aperture at any opening. 
Once locked, the fingers and grasped object form a stiff extension of the tool; the F/T sensor then measures only \textit{external} contact wrenches.
Unlike ratcheted locks and soft fingertips in previous work~\cite{ForceMimic}, the continuous lock avoids quantization and enables high gripping forces across object sizes, maintaining hold even under significant tensile loads.
Our software optionally filters any residual trigger-coupled components when only external contacts are desired for segmentation.

\subsubsection{ViTac Sensing for Contact-Rich Transitions}
Custom fingertip slots accept ViTac modules (Fig.~\ref{fig:gripper_structure} \textcircled{1}), which capture fine surface deformations (Fig.~\ref{fig:cable_mounting_process}). 
These tactile cues disambiguate visually similar states and thus
% —e.g., pulling a cable taut vs.\ seating it in a clip—
improve phase understanding and segmentation in contact-centric tasks. 
% Fingertips are modular and can be swapped when tactile input is unnecessary. 

\subsubsection{Drift-Free 6-DoF Pose Tracking}
A rigidly mounted Vive Tracker on the top cover provides accurate 6-DoF trajectories with negligible long-horizon drift, enabling a fixed transform to the tool center point (TCP). 
Compared with visual–inertial SLAM pipelines~\cite{umi}, this eliminates cumulative drift during extended demonstrations and yields stable trajectories for retargeting.

In summary, with a robot-compatible F/T stack with a \emph{continuous} mechanical lock, the device isolates \emph{external} interaction wrenches during demonstrations, which prior UMI-family designs lack. 
Integrated further with tactile and pose sensing, it is a compact, use-friendly and robot-compatible data collection tool which supports synchronized multi-modal
sensing and is especially suitable for capturing contact-rich interactions in long-horizon demonstrations.
Ergonomically, it supports single-handed operation: the center of mass sits near the palm, the trigger is indexed to the forefingers, and the lock is thumb-engaged with index release. 
Meanwhile, it preserves wrist–tool configuration which allows seamless transfer to real robots.
The overall structure is lightweight and largely composed of 3D-printed components, making it easy to manufacture, customize, and reproduce.

\begin{figure*}[thpb]
    \centering
    \includegraphics[width=0.95\textwidth]{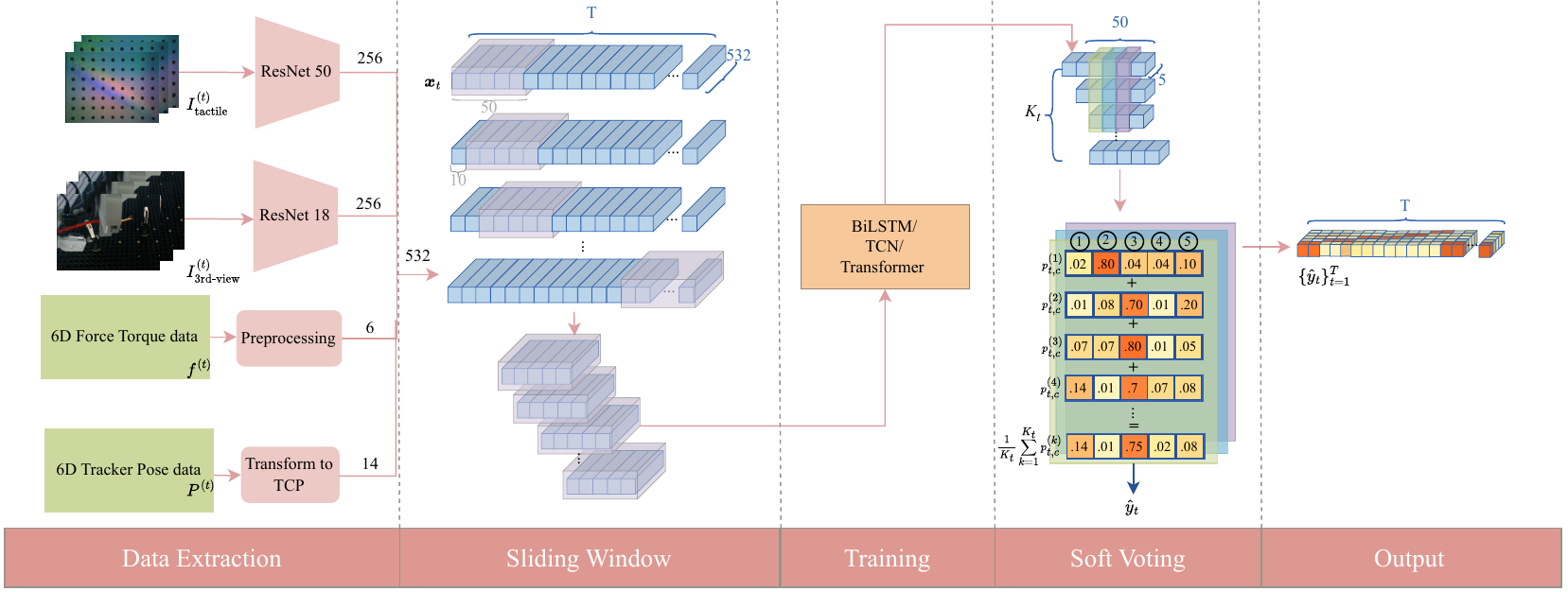}
    \vspace{-0.5em}
    \caption{\textbf{Pipeline of the proposed event segmentation algorithm.} 
    The framework consists of four main stages: 
    (a) \textbf{Data Extraction:} tactile images are processed by a ResNet50 to obtain 256-dimensional embeddings; third-person view images are processed by a ResNet18 with GroupNorm to obtain another 256-dimensional visual embedding; raw 6D F/T signals are preprocessed and 6D tracker pose are transformed to TCP.
    (b) \textbf{Sliding Window:} fused feature sequences of size $[T,532]$ are segmented into overlapping windows of length 50 with stride 10. 
    (c) \textbf{Training:} each window is fed into one of three sequence models (BiLSTM, TCN, or Transformer) to capture temporal dependencies and produce per-frame skill predictions.  
    (d) \textbf{soft voting inference:} overlapping predictions for the same frame are aggregated by averaging class probabilities and assigning the highest-probability label, restoring the original sequence length for final output.}
    \label{fig:algorithm_pipeline}
    \vspace{-1.0em}
\end{figure*}

\subsection{Event Segmentation}\label{sec:segmentation}
We aim to segment long-horizon manipulation demonstrations temporally into meaningful short-skill segments using rich multi-modal sensory feedback. 
% Unlike prior work\cite{LEMMo-Plan} that relies solely on tactile information, 
Our approach leverages synchronized \textit{tactile images, force-torque data, third-person view RGB images, and pose data} to detect skill transitions with high granularity.

\subsubsection{Network Architecture}

At each timestep, we fuse representation from tactile, RGB image, F/T and pose (Fig.~\ref{fig:algorithm_pipeline}(a)). 
For tactile, we feed raw ViTac frames to a ResNet-50 backbone pretrained on ImageNet, truncate the classification head, and add a linear layer to yield a 256-D embedding.
For the third-person views, inspired by Diffusion Policy~\cite{diffusion_policy}, we use a ResNet-18 with all BatchNorm replaced by GroupNorm, remove global average pooling and fully connected layers, apply a Spatial softmax to extract spatial attention features, and project to a 256-D embedding.
The raw 6-D F/T stream is preprocessed to remove operator trigger artifacts (\textit{pull}, \textit{lock}, \textit{release}) using a secondary segmentation model (see Section~\ref{subsec:F/T_preprocessing}), yielding an interaction-only signal (Fig.~\ref{fig:F/T_data_comparison}b).
The pose data obtained from trackers are transformed into the grippers’ tool center point (TCP). For each hand, we represent the TCP pose as a 7-D vector. Concatenating both left and right hand TCP poses yields a 14-D pose feature.
The resulting vectors are normalized and concatenated with the tactile and visual embeddings to form
\begin{equation}
\boldsymbol{x}_t=\big[\boldsymbol{I}_{\text{tactile}}^{(t)} \,\|\, 
\boldsymbol{I}_{\text{3rd-view}}^{(t)} \,\|\, 
\boldsymbol{f}^{(t)} \,\|\, 
\boldsymbol{P}^{(t)}\big] \in \mathbb{R}^{532} \text{,}
\end{equation}
where 
% $\boldsymbol{I}_{\text{tactile}}^{(t)} \in \mathbb{R}^{256}$, 
% $\boldsymbol{I}_{\text{3rd-view}}^{(t)} \in \mathbb{R}^{256}$, 
$\boldsymbol{I}_{(\cdot)}^{(t)} \in \mathbb{R}^{256}$, 
$\boldsymbol{f}^{(t)} \in \mathbb{R}^{6}$, and 
$\boldsymbol{P}^{(t)} \in \mathbb{R}^{14}$.

To capture temporal dependencies across multimodal sequences, we evaluate three different backbone architectures: 
% \begin{enumerate}
\begin{itemize}
\item a three-layer Bidirectional LSTM (BiLSTM) with hidden size 128 per direction. 
Taking the fused multi-modal representation $\boldsymbol{x}_t$ at each timestep as input, the BiLSTM produces a sequence of hidden states $\{\boldsymbol{h}_t\}_{t=1}^T$, which are subsequently passed through a dropout layer and a linear classifier to produce $\boldsymbol{z}_t \in \mathbb{R}^{C}$,
% \begin{equation}
% \boldsymbol{z}_t = W\boldsymbol{h}_t + \boldsymbol{b}, \quad 
% \boldsymbol{z}_t \in \mathbb{R}^{C},
% \label{eq:logits}
% \end{equation}
where $C$ denotes the number of skills.  
\item  a Temporal Convolutional Network (TCN) composed of stacked residual blocks with dilated 1D convolutions, GELU activations, and Layer Normalization. 
This architecture enlarges the receptive field and captures multiscale temporal patterns while maintaining efficient computation.  
\item a Transformer Encoder with learned linear input projection, sinusoidal positional encoding, and $L$ stacked encoder layers, each consisting of multi-head self-attention and feedforward sublayers.  
% shown in equation \eqref{eq:seq_mapping}. 
% \begin{equation}
% f: \mathbb{R}^{T \times 532} \;\rightarrow\; \mathbb{R}^{T \times C}
% \label{eq:seq_mapping}
% \end{equation}

% \end{enumerate}
\end{itemize}

All three architectures output frame-wise logits from sequence-to-sequence mapping $f: \mathbb{R}^{T \times 532} \;\rightarrow\; \mathbb{R}^{T \times C}$.

% Given a sequence of $T$ fused embeddings $\{\boldsymbol{x}_1, \dots, \boldsymbol{x}_T\}$, the BiLSTM outputs a sequence of hidden representations $\{\boldsymbol{h}_1, \dots, \boldsymbol{h}_T\}$.
% We attach a linear classification head to each hidden state to produce class logits,
% \begin{equation}
% \boldsymbol{z}_t = W\,\boldsymbol{h}_t + \boldsymbol{b}, \quad t=1,\dots,T, \label{eq:logits}
% \end{equation}
% and obtain class posteriors via the softmax,
% \begin{equation}
% \boldsymbol{y}_t = \operatorname{softmax}(\boldsymbol{z}_t), \label{eq:softmax}
% \end{equation}

\subsubsection{Sliding Window and Training Strategy}
In sequential tasks, there is usually a label imbalance in terms of number of frames.
Particularly, the \textit{idle} class where nothing happens are often over-represented.
To handle this imbalance and also to increase training efficiency, we adopt a sliding window to extract fixed-length overlapping subsequences from each demonstration(Fig.~\ref{fig:algorithm_pipeline}(b)).
This strategy both augments the dataset and improves the model's focus on meaningful skill transitions.

We use a window size of 50 and a stride of 10. 
To ensure that the model learns from active manipulation segments, we discard any window in which more than 80\% of the frames are labeled as \textit{idle}, enforcing a minimum action ratio of 20\%.
Each retained window is paired with its corresponding frame-wise label sequence and used to supervise the temporal model. Training is conducted on the long-horizon task using the Adam optimizer with standard learning rate scheduling. To prevent overfitting, we apply dropout after the BiLSTM layers and employ early stopping based on validation segmentation accuracy.

\subsubsection{Soft Voting for Inference}
After the BiLSTM model predicts frame-wise skill labels for each input window, we obtain multiple overlapping predictions for the same frame due to the use of a sliding window during both training and inference. Each frame in the original demonstration may appear in several windows, each providing an independent classification output.

To reconcile these overlapping predictions and restore the temporal alignment with the original sequence, we apply a soft voting strategy. Specifically, for each frame $t$ in the original sequence, we collect all predicted class probability vectors from the windows that include frame $t$.
We aggregate probabilities from all overlapping windows that include $t$. 
Specifically, let $K_t$ denote the number of such windows and $p^{(k)}_{t,c}$ the predicted probability of class $c$ from the $k$-th window.
We compute the averaged class probability and assign the final label by
\begin{equation}
\hat{y}_t = \arg\max_{c \in \{1,\dots,C\}} 
\frac{1}{K_t}\sum_{k=1}^{K_t} p^{(k)}_{t,c}, \label{eq:majority-vote}
\end{equation}
so that $\hat{y}_t$ corresponds to the class with the highest mean probability across windows. As shown in \eqref{eq:majority-vote}.

This soft aggregation method ensures temporal consistency, reduces prediction noise, and produces a smooth, frame-aligned label sequence for the entire demonstration. The final output preserves the original sequence length and provides a single semantic label per frame, suitable for downstream evaluation and visualization.

% \subsection{Use in Preprocessing: Trigger Action Filtering}\label{subsec:F/T_preprocessing}

% A unique downstream use case of our segmentation model is to \textbf{filter out trigger-related transitions} in force-torque signals. Since pulling, locking and releasing the trigger introduces internal forces that are not related to external interaction, we train a second instance of the same segmentation architecture to explicitly classify these intervals.  
% Instead of simply removing them, we replace the corresponding force-torque values with Gaussian noise whose mean and standard deviation are estimated from the first 20 frames of each sequence, where no trigger actions occur. This noise injection preserves the natural statistical characteristics of the signal, as shown in Fig.~\ref{fig:F/T_data_comparison}(b), while effectively removing internal actuation effects.  
% The resulting interaction-only force-torque stream maintains temporal continuity and is better suited for downstream robotic learning and control.

\subsection{F/T Data Preprocessing}\label{subsec:F/T_preprocessing}
\begin{figure}[t!]
    \centering  
    \includegraphics[width=0.95\columnwidth]{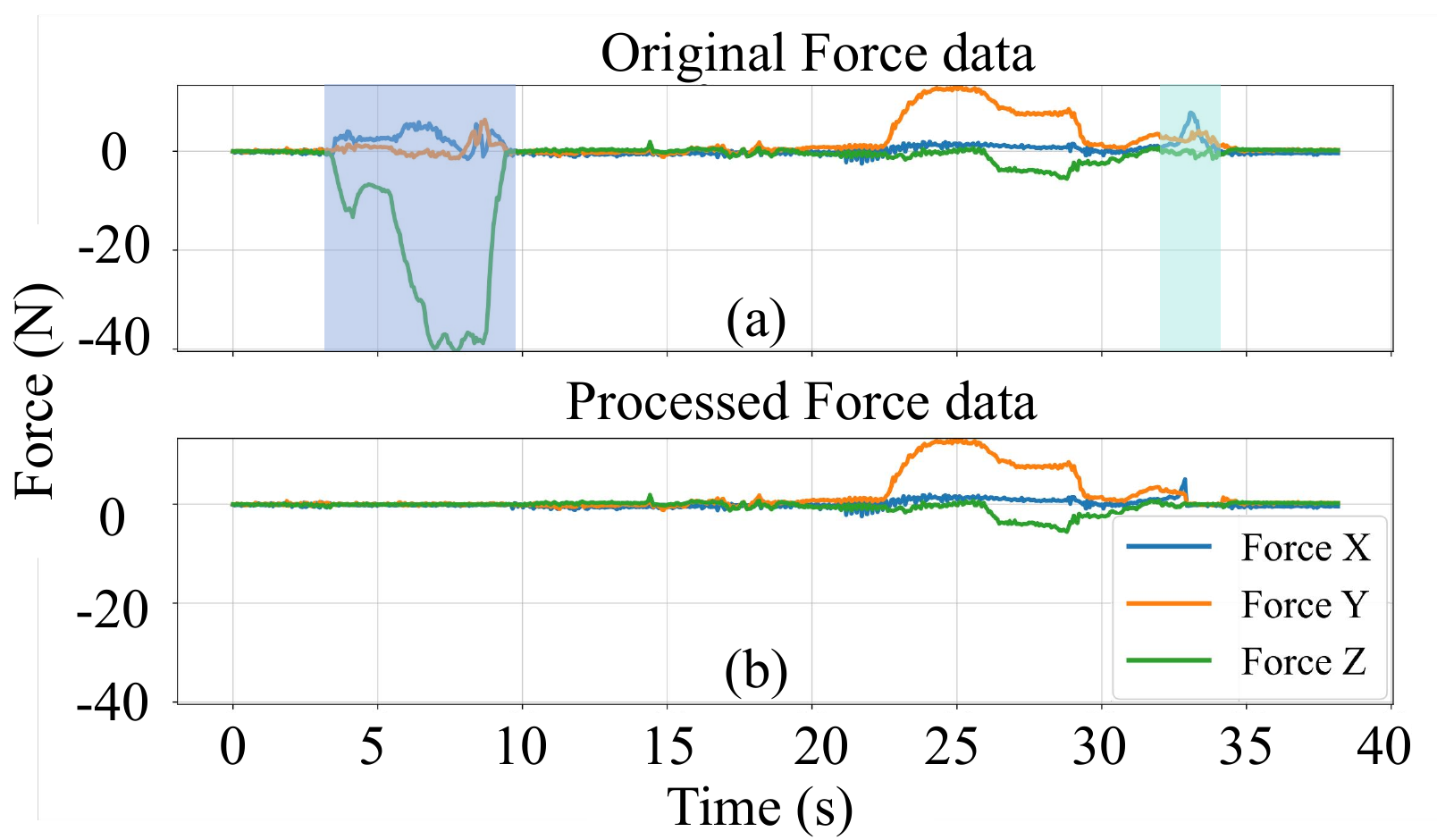}
    \caption{\textbf{Comparison of original and processed Force/Torque (F/T) data.} 
    (a) \textbf{Original F/T data:} The recorded signals reveal distinct operator-induced events, namely pulling and locking the trigger (blue-shaded region) followed by releasing it (green-shaded region), both of which cause significant disturbances in the measured force and torque signals.  
    (b) \textbf{Processed F/T data:} The influence of these three actions has been removed, yielding clean signals that can be directly utilized by the robot without the undesired human-induced artifacts.}
    \label{fig:F/T_data_comparison}
    \vspace{-1em}
\end{figure}
In this section, we describe two preprocessing steps applied to raw force-torque (F/T) signals before they are used for segmentation.

\subsubsection{Trigger Action Filtering}
A unique downstream use case of our segmentation model is to \textbf{filter out trigger-related transitions} in force-torque signals. 
As shown in Fig.~\ref{fig:F/T_data_comparison}(a), with help of the continuous locking mechanism, the collected force and torque signals drop sharply to near zero right after locking, which indicates that internal actuation effects have been eliminated and resembles the measurement of wrist-mounted F/T sensors on robots.
Nevertheless, there are still temporary interferences when the operator pulls the trigger, locks it or releases it.
% Since pulling, locking and releasing the trigger introduces internal forces that are not related to external interaction, 
To filter these interferences out, we train a second instance of the same segmentation architecture to explicitly classify these intervals.  
We then replace the identified interference intervals with Gaussian noise whose mean and standard deviation estimated from the first 20 frames of each sequence, where no trigger actions occur.
% Instead of simply removing them, we replace the corresponding force-torque values with Gaussian noise whose mean and standard deviation are estimated from the first 20 frames of each sequence, where no trigger actions occur. 
This noise injection preserves the natural statistical characteristics of the signal, as shown in Fig.~\ref{fig:F/T_data_comparison}(b), while effectively removing internal actuation effects.  
The resulting interaction-only force-torque stream maintains temporal continuity and is better suited for downstream robotic learning and control.

\subsubsection{Gripper-to-Robot F/T Data Mapping}
% Another important preprocessing step is to map the F/T readings measured at our custom gripper back to an equivalent representation at the robot end-effector. This is crucial because the relative placement of the wrist-mounted F/T sensor with respect to the tool center point (TCP) differs across robot platforms. Our gripper design follows the convention that the $z$-axis of the F/T sensor is aligned with the TCP $z$-axis, such that the relative transformation only involves a rotation about $z$ and a translation along the $z$-direction.  
To adapt the gripper-collected F/T data for use across different robot platforms, we map the measured wrench from the gripper F/T sensor frame $\{\mathrm{G}\}$ to the robot end-effector frame $\{\mathrm{R}\}$. The mapping is given by

\begin{equation}
\mathbf{F}_R = R_{GR}\mathbf{F}_G,
\label{eq:force_mapping}
\end{equation}
\begin{equation}
\boldsymbol{\tau}_R = R_{GR}\boldsymbol{\tau}_G + \mathbf{r}_{GR} \times \left(R_{GR}\mathbf{F}_G\right),
\label{eq:torque_mapping}
\end{equation}
where 
$\mathbf{F}_G \in \mathbb{R}^3$ and $\boldsymbol{\tau}_G \in \mathbb{R}^3$ 
denote the force and torque measured in $\{\mathrm{G}\}$, 
$\mathbf{F}_R \in \mathbb{R}^3$ and $\boldsymbol{\tau}_R \in \mathbb{R}^3$ 
are the corresponding quantities expressed in $\{\mathrm{R}\}$, 
$R_{GR} \in SO(3)$ is the rotation matrix (expressed in the TCP coordinate system) that transforms vectors from the gripper-mounted F/T sensor frame to the robot-mounted F/T sensor frame, and 
$\mathbf{r}_{GR} \in \mathbb{R}^3$ is the displacement vector (expressed in the TCP coordinate system) from the measurement center $O_G$ of the gripper-mounted F/T sensor to the measurement center $O_R$ of the robot-mounted F/T sensor.  

Equations~\eqref{eq:force_mapping} and \eqref{eq:torque_mapping} follow the standard wrench transformation under a change of reference frame, and ensure that the interaction forces measured at the gripper are consistently expressed in the robot end-effector frame. This unified representation enables seamless transfer of the gripper-collected F/T data across robot platforms with varying wrist sensor and TCP configurations.

\section{EXPERIMENTS}
\begin{figure*}[thpb]
    \centering
    \includegraphics[width=0.95\textwidth]{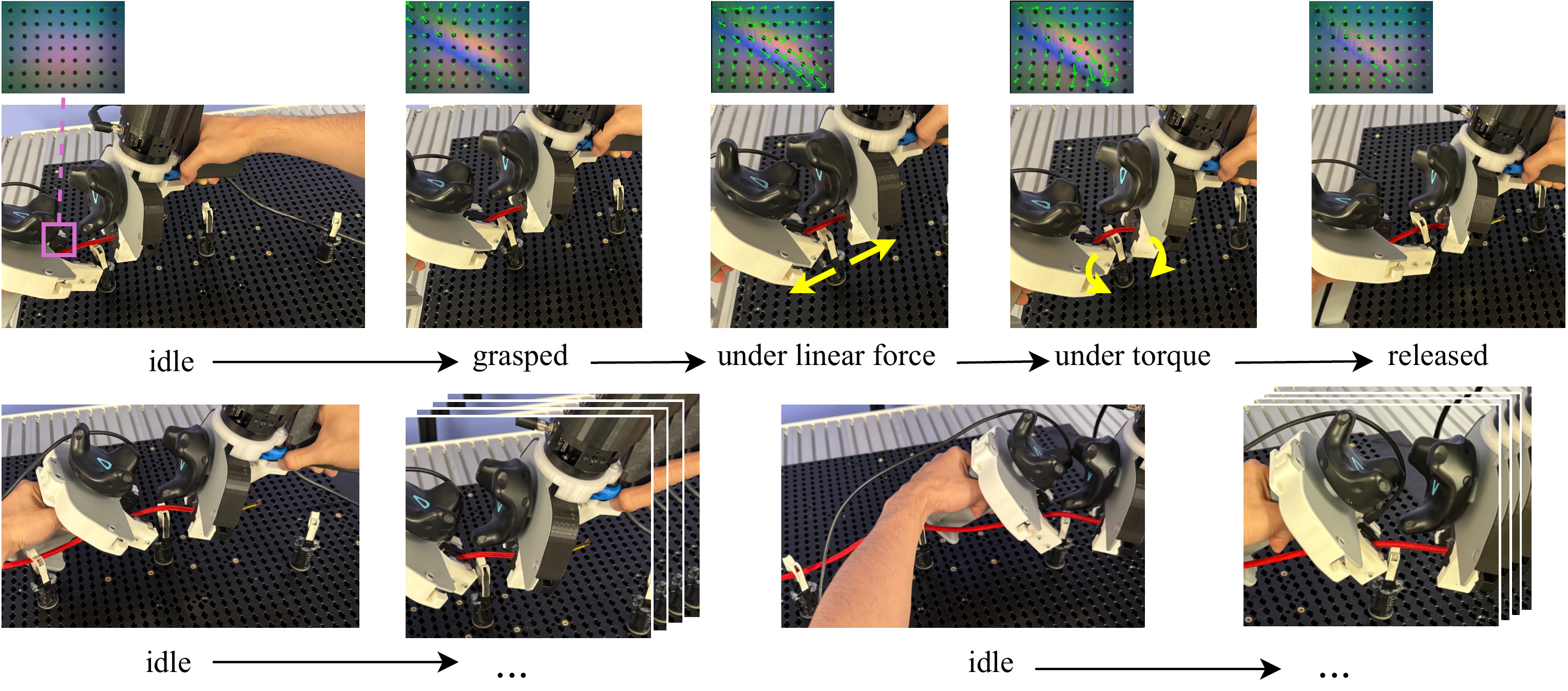}
    \vspace{-0.5em}
    \caption{\textbf{Overview of the cable mounting task process.} 
    Using the proposed \textbf{TacUMI}, the cable is sequentially inserted into three U-type clips with different orientations. 
    }
    \label{fig:cable_mounting_process}
    % \vspace{1.5em}
\end{figure*}

In this section, we perform a cable mounting experiment to validate both the correctness and efficiency of multimodal data collection using our proposed handheld gripper, and to demonstrate the effectiveness of the event segmentation algorithm. 

\subsection{Experimental Setup}
\subsubsection{Hardware and Sensors}
We conduct experiments using our custom-designed handheld tools, whose multi-modal data collection system consists of four sensors: a {Bota Systems SensONE} 6D force-torque sensor, a {GelSight Mini} ViTac sensor, an {HTC Vive Tracker} for 6-DoF pose tracking, and a fixed-position {RealSense D435i camera} for third-person visual observation.
These sensors operate at different frequencies: the force-torque sensor at 1000\,Hz, the ViTac sensor (with online processing) at 16.67\,Hz, the Vive Tracker at 60\,Hz, and the RGB-D camera also at 60\,Hz.
To ensure temporal alignment across modalities, we down-sample all sensor streams to a common frequency of 16.67\,Hz, matching that of the tactile sensor, the lowest among all devices. 

% For dual-arm manipulation tasks, we could distribute collection of different modalities to different hands.
% For example, in cable manipulation, we want to capture force-torque measurements from the right hand which maintains grasping for the whole time, and capture ViTac streaming from the left hand which repeatedly grasping and releasing.
% In such a case, the right-hand gripper is equipped with a F/T sensor and  the left-hand gripper with a ViTac sensor, and each hand with a Vive Tracker.

For dual-arm manipulation tasks, we distribute different sensing modalities to different hands.  
For example, in the cable mounting task(shown in Fig.~\ref{fig:cable_mounting_process}), the right-hand TacUMI is equipped with a force–torque sensor and a Vive Tracker to maintain stable grasping and guide the cable along its path, while the left-hand gripper carries a ViTac sensor and a Vive Tracker to perform fine manipulations, including tensioning, positioning, and inserting the cable into clips.  
During each trial, the operator progresses along the clip sequence, coordinating both grippers to maintain cable stability while executing precise insertions.  
Variations in clip position and orientation introduce non-repetitive transitions between subtasks, providing a challenging benchmark for multimodal task segmentation and validating the robustness of our proposed approach.  

To validate that the multimodal demonstrations collected with our handheld gripper can be effectively transferred to a robot platform, we additionally collect a test dataset from teleopration, where we employ \textit{sigma.7} haptic devices to control two \textit{Franka Emika} robots in a master–slave configuration.  
The teleoperation setup mirrors the handheld gripper setup:  
the guiding arm is equipped with a wrist-mounted F/T sensor,  
and the manipulating arm is equipped with a fingertip-mounted tactile sensor, so it captures the the same modalities (tactile, force-torque, and pose from proprioceptive data) as the handheld device.
We then evaluate on the teleopration dataset whether the segmentation models trained exclusively on handheld-gripper data could generalize well to robot-collected data.
% In this way, the teleoperation experiments provide a principled means to verify the cross-platform applicability of our data collection device and the robustness of the proposed segmentation framework.
% This configuration ensures that both the handheld and the teleoperation setups capture the same multimodal signals: tactile, force-torque, and pose, while differing only in the data collection interface.
\begin{figure}[thpb]
    \centering  
    \includegraphics[width=0.95\columnwidth]{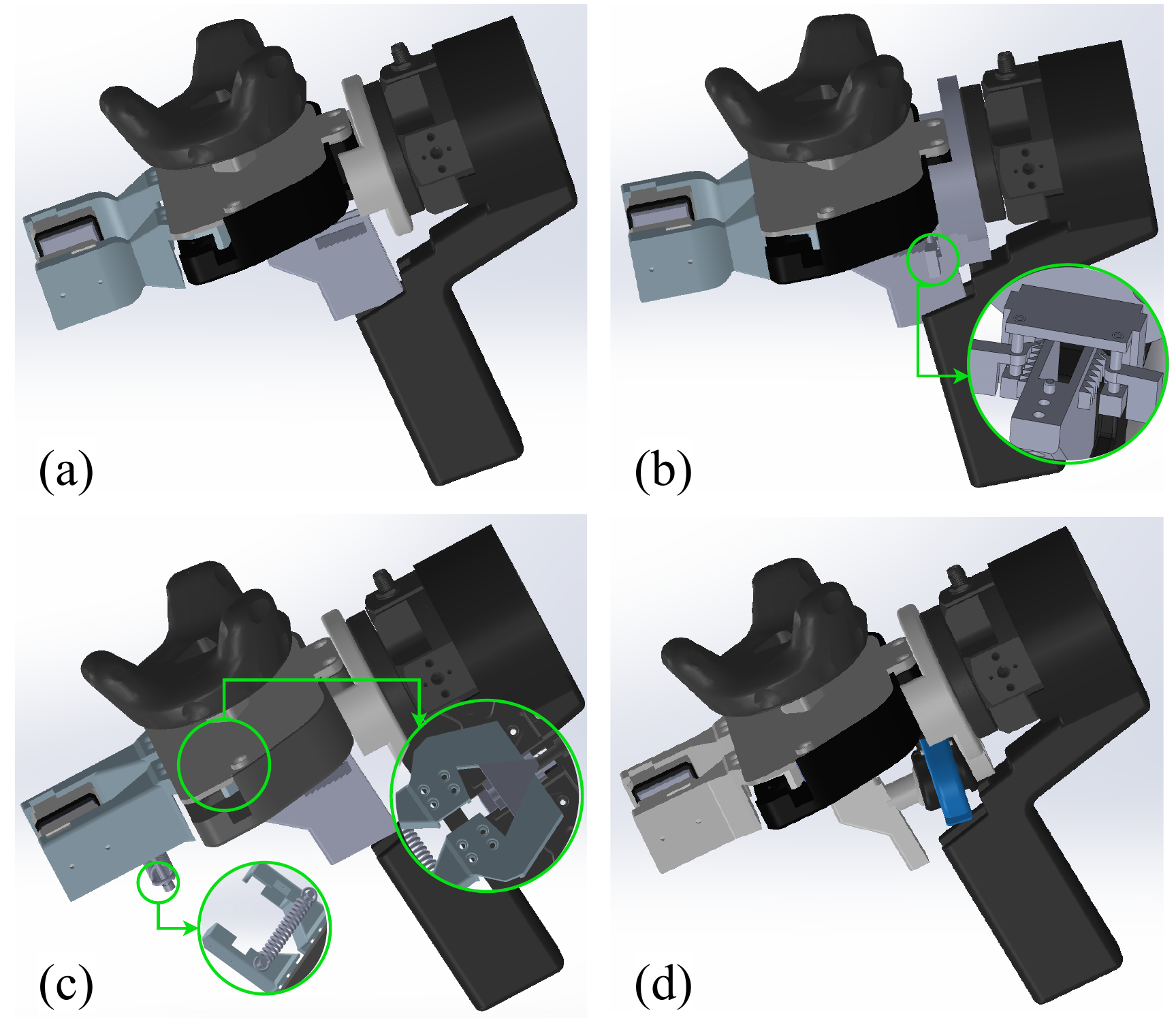}
    \caption{\textbf{Comparison of four gripper designs.} 
    (a) Design without a locking mechanism: the fingertips are closed by pulling the trigger backward.  
    (b) Design with a ratchet-based locking mechanism: once locked, the trigger can be released by pushing forward the two levers showed in the green circle.  
    (c) Design with tension spring-driven fingertips: the fingertips are normally closed by a tension spring (showed in the green circle) and open when the trigger is pulled backward.  
    (d) Design with a continuous locking mechanism: the proposed design introduced in detail in previous figure, enabling stable and adjustable locking at arbitrary grasping width.}
    \label{fig:different_designs}
    \vspace{-2.0em}
\end{figure}

\subsection{Data Collection Correctness and Efficiency}

To evaluate the correctness of F/T data collected by different gripper designs, we compared four variants as illustrated in Fig.~\ref{fig:different_designs}.  
The first design (Fig.~\ref{fig:different_designs}(a)) places the F/T sensor between the handle and the gripper body, but does not include any locking mechanism, so the jaws are closed only by continuously pulling the trigger.  
The second design (Fig.~\ref{fig:different_designs}(b)) adds a ratchet-based locking mechanism, similar in principle to ForceMimic~\cite{ForceMimic}, allowing the trigger to be locked and released via mechanical levers.  
The third design (Fig.~\ref{fig:different_designs}(c)) employs a reverse actuation structure with a tension spring mounted under the fingertips, keeping them normally closed, pulling the trigger opens the fingertips against the spring force.  
The fourth design (Fig.~\ref{fig:different_designs}(d)) corresponds to our proposed continuous locking mechanism described in Section~\ref{sec:hardware_design}, which ensures stable grasping during demonstrations.

We used each design to collect multimodal data during the cable mounting task and mapped the measured F/T signals into the robot end-effector frame using Equations~\eqref{eq:force_mapping} and \eqref{eq:torque_mapping}. 
% \begin{figure*}[thpb]
%     \centering
%     \includegraphics[width=0.95\textwidth]{figures/5_Force (cropped) (pdfresizer.com).pdf}
%     \vspace{-0.5em}
%     \caption{\textbf{Comparison of force measurements in the cable mounting task across different gripper designs and teleoperated robot.} 
%     (a) Gripper without locking mechanism. 
%     (b) Gripper with ratchet-based locking mechanism. 
%     (c) Gripper with tension spring-based mechanism. 
%     (d) Proposed \textbf{TacUMI} gripper integrating tactile and force sensing. 
%     (e) Teleoperated robot with wrist-mounted FT sensor.}
%     \label{fig:forces_comparison}
%     % \vspace{1.5em}
% \end{figure*}
\begin{figure}[thpb]
    \centering  \includegraphics[width=0.95\columnwidth]{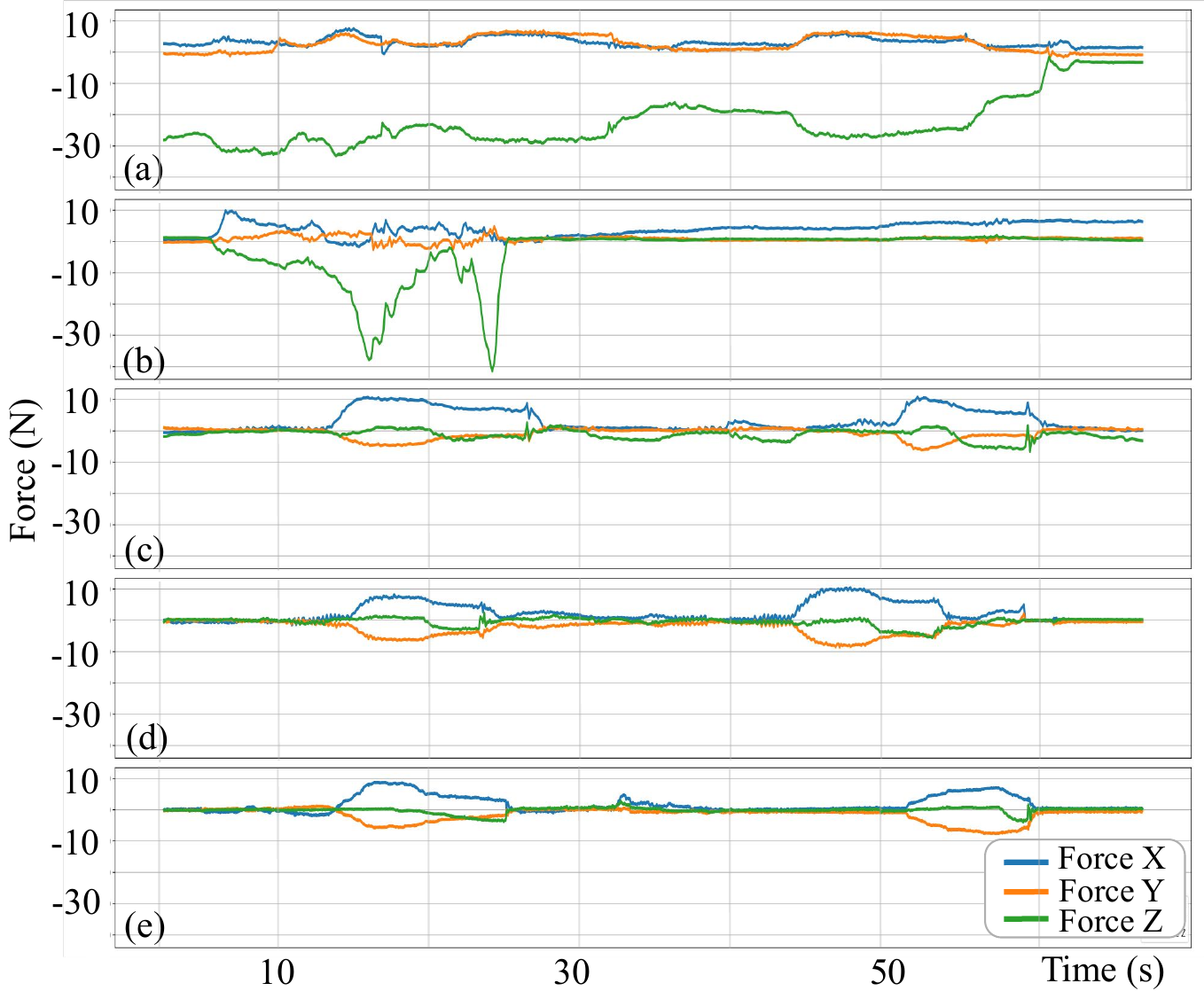}
    \caption{\textbf{Comparison of force measurements in the cable mounting task across different gripper designs and teleoperated robot.} 
    (a) Gripper without locking mechanism. 
    (b) Gripper with ratchet-based locking mechanism. 
    (c) Gripper with tension spring-based mechanism. 
    (d) Proposed \textbf{TacUMI} gripper integrating tactile and force sensing. 
    (e) Teleoperated robot with wrist-mounted FT sensor.}
    \label{fig:forces_comparison}
    \vspace{-2em}
\end{figure}
As shown in Fig.~\ref{fig:forces_comparison}(a), the first design successfully detected F/T data, however, since the operator had to continuously hold the trigger, an additional non-constant actuation force was introduced, which could not be removed through preprocessing. 
In Fig.~\ref{fig:forces_comparison}(b), the second design with a ratchet-based locking mechanism exhibited discontinuous locking of the grasping width. As a result, the gripper failed to securely hold the cable, causing it to slip out during tightening and leading to unsuccessful task execution. 
In contrast, both Fig.~\ref{fig:forces_comparison}(c) and Fig.~\ref{fig:forces_comparison}(d) show F/T data that are consistent with those collected from the teleoperation in Fig.~\ref{fig:forces_comparison}(e). Nevertheless, the spring-based design in Fig.~\ref{fig:different_designs}(c) lacks generalizability: due to the stiffness of the tension spring, it cannot grasp cables with larger diameters or objects requiring wider fingertip openings. 
In contrast, our final design TacUMI shown in Fig.~\ref{fig:different_designs}(d) not only enables grasping across a wide range of grasping widths but also produces F/T data, that (after preprocessing) closely match those obtained from teleoperated robot with a wrist-mounted F/T sensor. The minor deviations are attributable to execution differences between handheld and teleoperated demonstrations. This confirms that the TacUMI design achieves correct and robot-consumable F/T data collection.

Beyond correctness, we also compare the efficiency of teleoperation with our handheld gripper design. In each trial of the cable mounting task, the cable had to be sequentially inserted into three clips arranged in different orientations. 
% Each successful insertion was characterized by four discrete cable states: \textit{grasped}, \textit{under linear force}, \textit{under torque}, and \textit{released}. 
We measured the total task completion time across multiple trials for both methods. Results show that the teleoperation approach required an average of \textbf{4 minutes} per trial, whereas our handheld gripper completed the same task in only \textbf{1 minute 10 seconds} on average. This demonstrates a substantial improvement in data collection speed, enabling faster acquisition of high-quality multimodal datasets for long-horizon task segmentation.

\subsection{Ablation on Model Architectures and Input Modalities}
\subsubsection{Evaluation Metrics}
We evaluate our event segmentation models using the following metrics:

\begin{itemize}
    \item \textbf{Frame-wise Accuracy:}  
    The proportion of correctly predicted labels over all frames in the restored sequence after applying the soft voting strategy.  
    
    \item \textbf{F1 Score per Class:}  
    Harmonic mean of precision and recall for each skill class.
    
    % \item \textbf{Training Time:}  
    % Total time for model convergence.
\end{itemize}

We conduct an extensive ablation study to analyze the impact of model architectures and input modalities on segmentation performance.  
For architectures, we compare BiLSTM, TCN, and Transformer encoder introduced in Sec.~\ref{sec:segmentation}.  
For input modalities, we evaluate single-modality settings,  
dual-modality combinations,  
three-modality combinations,  
and the full multimodal configuration(Table~\ref{tab:ablation_arch_modal_gripper}).  

% Table~\ref{tab:ablation_arch_modal_gripper} summarizes results when testing on datasets collected with the handheld gripper.

\begin{table}[h]
\footnotesize
% \caption{Frame-wise accuracy (\%) for different model architectures and input modalities (Test set: Gripper data).}
\caption{Frame-wise accuracy (\%) on TacUMI data.}
\begin{center}
\resizebox{0.48\textwidth}{!}{
\begin{tabular}{|l|c|c|c|}
\hline
\textbf{Input Modality} & \textbf{\textit{BiLSTM}} & \textbf{\textit{TCN}} & \textbf{\textit{Transformer}} \\
\hline
Camera only & 0.7608 & 0.7217 & 0.3180 \\
\hline
Camera + Tactile & 0.9076 & 0.8880 & 0.6765 \\
\hline
Camera + F/T & 0.8632 & 0.8325 & 0.6645 \\
\hline
Camera + Pose & 0.8165 & 0.7675 & 0.4242 \\
\hline
Camera + Tactile + F/T & 0.9359 & 0.9051 & 0.7459 \\
\hline
Camera + Tactile + F/T + Pose & \textbf{0.9402} & 0.8945 & 0.7596 \\
\hline
\end{tabular}}
\vspace{-1.0em}
\label{tab:ablation_arch_modal_gripper}
\end{center}
\end{table}

We first analyze the frame-wise accuracy results.  
BiLSTM consistently outperforms TCN and Transformer, as it can exploit both past and future context to accurately detect event boundaries and exhibits greater robustness to limited data scale (30k frames) and sensor noise, whereas the Transformer encoder requires larger datasets and is more sensitive to noise, and the TCN lacks the ability to access bidirectional temporal context.   
In terms of input modalities, models relying solely on third-person visual input perform worst, often failing to identify subtle skill transitions.  
Adding tactile or F/T signals substantially boosts accuracy, while TCP pose alone provides only marginal gains.  
The best performance is obtained when combining all four modalities, with BiLSTM achieving the highest overall accuracy.

% \begin{figure}[thpb]
%     \centering  
%     \includegraphics[width=0.95\columnwidth]{figures/training_time.png}
%     \caption{\textbf{Comparison of training times across input modalities and model architectures.} 
%     Average training time per epoch is reported for three segmentation architectures (BiLSTM, TCN, Transformer) under different input modality settings.}
%     \label{fig:training_time}
%     \vspace{-1.0em}
% \end{figure}

% Next, we examine the training efficiency of different architectures (Fig.~\ref{fig:training_time}).  
% For the same input modality, Transformer requires significantly longer training time than BiLSTM and TCN, due to the $O(T^{2})$ complexity of its self-attention mechanism.  
% This overhead becomes prohibitive for our long-sequence cable mounting task.  
% Training time increases as more modalities are added, but the growth is relatively modest: for BiLSTM and TCN, the full multimodal configuration requires only about 30 additional seconds compared to single visual input, yet yields an accuracy improvement of nearly 20\%.  
% These results highlight that incorporating multimodal signals is highly beneficial for event segmentation, and that our proposed architecture in particular offers the best balance between accuracy and efficiency.  

\begin{figure}[thpb]
    \centering  
    \includegraphics[width=0.95\columnwidth]{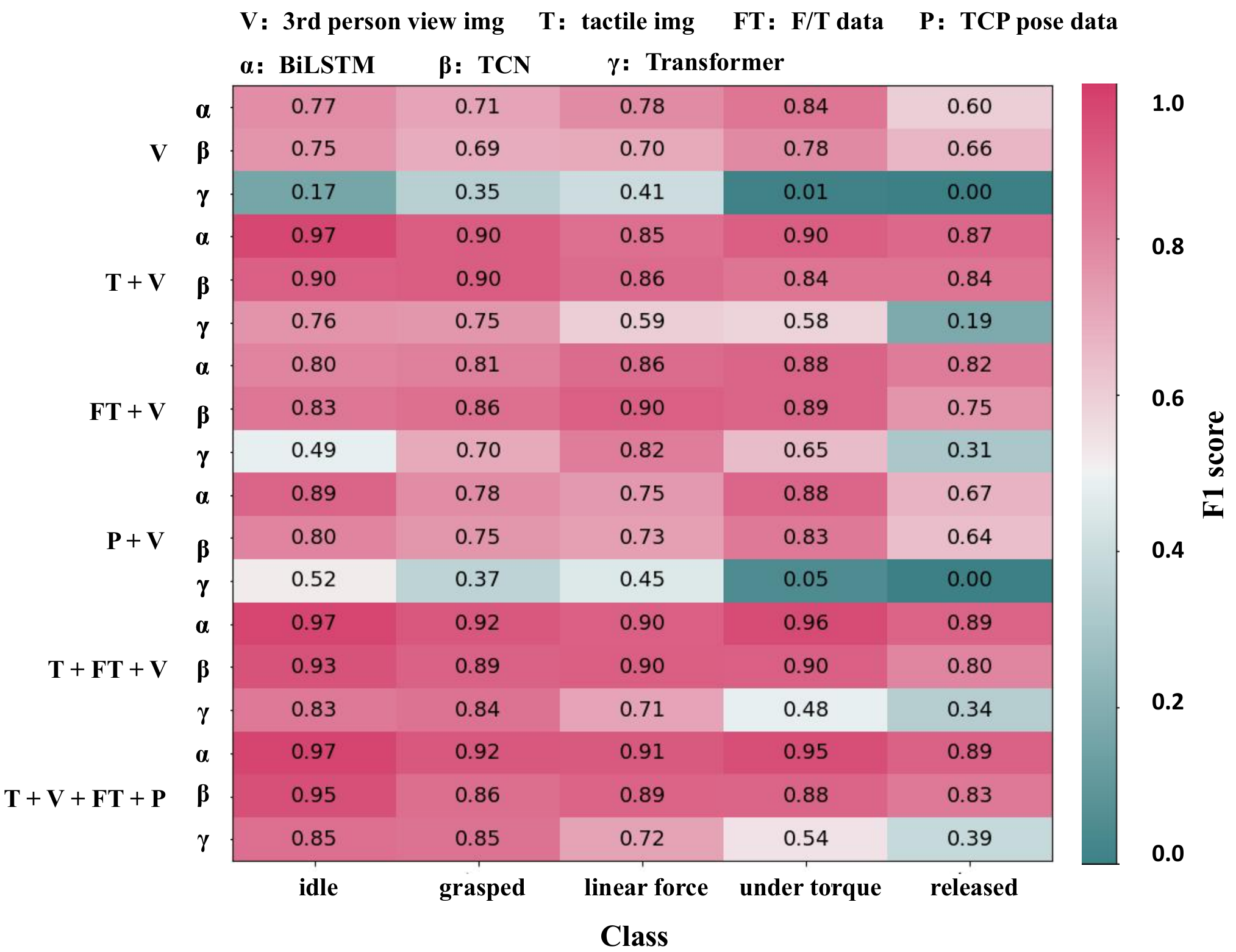}
    \caption{\textbf{Per-class segmentation performance across input modalities and model architectures.} 
    Heatmap of average F1-scores for each action class (\textit{idle, grasped, under a linear force, under torque, released}) across all modality–architecture combinations.  
    Higher values (towards red) indicate better class-wise segmentation accuracy.}
    \label{fig:f1_heatmap}
\end{figure}
Finally, we analyze class-wise segmentation performance (Fig.~\ref{fig:f1_heatmap}).  
For \textit{grasped} and \textit{idle} phases, most models achieve high F1 scores, as these skills involve clear and distinguishable signals.  
However, performance drops notably for the \textit{released} and \textit{under torque} classes, as the transitions into these states are subtle and less distinguishable from neighboring phases.  
The F1 score trends are consistent with the frame-wise accuracy results: performance improves as more modalities are incorporated, with BiLSTM consistently outperforming TCN and Transformer.  
Nevertheless, the improvement from three modalities (\textit{third-person view + tactile + F/T}) to the full multimodal configuration (adding TCP pose) is negligible.  
This outcome is expected for our cable mounting task, since TCP pose remains nearly unchanged during critical phases such as grasping, stretching the cable (\textit{under linear force}), and inserting it into the clip (\textit{under torque}), and thus provides little additional information for segmentation.  
Among all events, the \textit{released} class shows the lowest F1 score.  
This is primarily due to its extremely short duration, typically only 2–5 frames, so misclassifying even one or two frames leads to a sharp drop in F1 score.  
  
In particular, BiLSTM with the full multimodal input yields balanced performance across all classes, demonstrating that complementary sensing modalities are critical for capturing fine-grained manipulation events.

\subsection{Cross-Platform Validation on Robot Data}
To verify that multimodal demonstrations collected with the handheld gripper can be reliably transferred to robot platforms, we additionally evaluate the trained segmentation models on teleoperation-collected datasets. Specifically, all models are trained exclusively on \textbf{TacUMI} collected data but tested on robot-side demonstrations recorded via teleoperation. 

% Table~\ref{tab:ablation_arch_modal_robot} reports the frame-wise accuracy for different modality and architecture combinations on the teleoperation test set. We observe that the overall accuracy is slightly lower than that obtained with gripper-collected test data (Table~\ref{tab:ablation_arch_modal_gripper}), which can be attributed to differences in operator motion and execution dynamics between handheld and robot teleoperation. Importantly, the relative ranking of modalities remains unchanged: models that integrate tactile and F/T information consistently outperform those relying on vision alone, and BiLSTM achieves the highest accuracy across all input settings. 

As shown in Table~\ref{tab:ablation_arch_modal_robot}, when using only third-person camera input, the segmentation performance is very poor.
This is expected, since the test set consists of robot end-effector manipulations, whereas the training and validation sets contain demonstrations with the handheld TacUMI, leading to a substantial domain gap between the two embodiments.
Similar to the trends observed in Table~\ref{tab:ablation_arch_modal_gripper}, adding tactile or F/T signals significantly improves accuracy; however, the results remain 10–20\% lower than those achieved with gripper-collected test sets. This gap can be explained by the higher difficulty of controlling two robots via \textit{sigma.7} haptic devices, which introduces additional noise and less stable demonstrations compared to handheld execution. 
Importantly, when tactile and FT modalities are combined, the accuracy nearly matches that of the gripper test set, further confirming the critical role of multimodal fusion in achieving robust event segmentation across platforms.

\begin{table}[h]
\footnotesize
% \caption{Frame-wise accuracy (\%) for different model architectures and input modalities (Test set: Teleoperation robot data).}
\caption{Frame-wise accuracy (\%) on robot data}
\begin{center}
\resizebox{0.48\textwidth}{!}{
\begin{tabular}{|l|c|c|c|}
\hline
\textbf{Input Modality} & \textbf{\textit{BiLSTM}} & \textbf{\textit{TCN}} & \textbf{\textit{Transformer}} \\
\hline
Camera only & 0.2288 & 0.1820 & 0.2227 \\
\hline
Camera + Tactile & 0.7474 & 0.6002 & 0.5351 \\
\hline
Camera + F/T & 0.6611 & 0.6366 & 0.4126 \\
\hline
Camera + Pose & 0.4694 & 0.4636 & 0.2256 \\
\hline
Camera + Tactile + F/T & 0.9155 & 0.8793 & 0.8092 \\
\hline
Camera + Tactile + F/T + Pose & \textbf{0.9104} & 0.7262 & 0.7796 \\
\hline
\end{tabular}}
\label{tab:ablation_arch_modal_robot}
\end{center}
\vspace{-1.0em}
\end{table}

In conclusion, these results confirm that data collected with TacUMI not only supports accurate event segmentation in handheld demonstrations, but also transfers effectively to robot-executed tasks. This cross-platform generalization highlights the correctness of our data collection design and its suitability for downstream robot learning.

\section{CONCLUSION}
We presented TacUMI, a next-generation handheld data collection interface that extends the Universal Manipulation Interface family with multimodal sensing and task segmentation capabilities. 
By integrating fingertip tactile sensing, a wrist-mounted force–torque sensor, and drift-free 6-DoF pose tracking into a robot-compatible design, our system enables the acquisition of high-quality multimodal demonstrations for contact-rich, long-horizon tasks. 
The proposed continuous locking mechanism ensures clean force measurements and stable grasps.
On the algorithmic side, we introduced a sequence modeling framework that learns from collected multimodal-data to segment long-horizon tasks into short reusable skills. Experiments on cable mounting show that our approach achieves accurate segmentation, and transfers reliably from handheld demonstrations to robot-executed tasks.

Although the results validate both the hardware design and the segmentation framework, the potential of TacUMI for other tasks has not yet been revealed. First, we aim to broaden the evaluation to a wider range of contact-rich, long-horizon manipulation tasks in order to further establish the generalizability of TacUMI. Second, beyond task segmentation, TacUMI can serve as a platform for multimodal imitation learning, enabling robots to acquire executable skills directly from the segmented demonstrations. Together, these extensions would advance the utility of TacUMI from reliable data collection and segmentation toward end-to-end robot learning in complex manipulation domains.

\bibliographystyle{IEEEtran}
\bibliography{reference}

\end{document}